\pdfoutput=1 

\documentclass[twoside,11pt]{article}

\usepackage{blindtext}

\usepackage[preprint]{jmlr2e}

\usepackage{amsmath}
\usepackage{booktabs}
\usepackage{pgf}
\usepackage{import}
\usepackage{tikz,tkz-euclide,pdftexcmds,calc}
\usetikzlibrary{arrows,shapes.geometric}
\usetikzlibrary{positioning}
\usetikzlibrary{automata}
\usetikzlibrary{arrows,shapes,calc}
\usepackage{circuitikz}
\usepackage{algorithmic}
\usepackage[linesnumbered,ruled,noend]{algorithm2e}
\usepackage{bbm}
\usepackage{multirow}

\usepackage{amssymb}
\usepackage{pifont}
\usepackage{bm} 
\usepackage{makecell} 
\definecolor{ao(english)}{rgb}{0.0, 0.5, 0.0}
\newcommand{\cmark}{\color{ao(english)} \ding{51}}%
\newcommand{\xmark}{\color{red} \ding{55}}%

\hypersetup{hidelinks}

\newcommand{\libraryname}{Torchhd}

\usepackage{lastpage}
\jmlrheading{24}{2023}{1-\pageref{LastPage}}{3/23}{6/23}{23-0300}{Mike Heddes, Igor Nunes, Pere Vergés, Denis Kleyko, Danny Abraham, Tony Givargis, Alexandru Nicolau, and Alexander Veidenbaum}
\ShortHeadings{\libraryname{}}{Heddes, Nunes, Vergés, Kleyko, Abraham, Givargis, Nicolau, and Veidenbaum}

\firstpageno{1}

\begin{document}

\title{\libraryname{}: An Open Source Python Library to Support Research on Hyperdimensional~Computing and Vector~Symbolic~Architectures}

\author{\name Mike Heddes\textsuperscript{\normalfont{1}} \email mheddes@uci.edu
    \AND
    \name Igor Nunes\textsuperscript{\normalfont{1}} \email igord@uci.edu
    \AND
    \name Pere Vergés\textsuperscript{\normalfont{1}} \email pvergesb@uci.edu
    \AND
    \name Denis Kleyko\textsuperscript{\normalfont{2}} \email denis.kleyko@ri.se
    \AND
    \name Danny Abraham\textsuperscript{\normalfont{1}} \email dannya1@uci.edu
    \AND
    \name Tony Givargis\textsuperscript{\normalfont{1}} \email givargis@uci.edu
    \AND
    \name Alexandru Nicolau\textsuperscript{\normalfont{1}} \email nicolau@ics.uci.edu
    \AND
    \name Alexander Veidenbaum\textsuperscript{\normalfont{1}} \email alexv@ics.uci.edu
    \AND
   \textsuperscript{\normalfont{1}} \addr Department of Computer Science, University of California, Irvine, CA 92617, USA
   \AND
   \textsuperscript{\normalfont{2}} \addr
   Intelligent Systems Lab, Research Institutes of Sweden, 164 40 Kista, Sweden
}

\editor{Sebastian Schelter}

\maketitle

\begin{abstract}
Hyperdimensional computing (HD), also known as vector symbolic architectures (VSA), is a framework for computing with distributed representations by exploiting properties of random high-dimensional vector spaces. 
The commitment of the scientific community to aggregate and disseminate research in this particularly multidisciplinary area has been fundamental for its advancement. 
Joining these efforts, we present \libraryname{}, a high-performance open source Python library for HD/VSA. 
\libraryname{} seeks to make HD/VSA more accessible and serves as an efficient foundation for further research and application development.
The easy-to-use library builds on top of PyTorch and features state-of-the-art HD/VSA functionality, clear documentation, and implementation examples from well-known publications. Comparing publicly available code with their corresponding \libraryname{} implementation shows that experiments can run up to 100$\times$ faster. \libraryname{} is available at:
\url{https://github.com/hyperdimensional-computing/torchhd}.
\end{abstract}


\section{Introduction}
\label{sec:introduction}

Hyperdimensional computing (HD)~\citep{kanerva2009hyperdimensional}, also known as vector symbolic architectures (VSA)~\citep{gayler2003vector}, is a computing framework capable of forming compositional distributed representations. 
It originated within artificial intelligence and cognitive science~\citep{SmolenskyTensor1990, kussul1991associative, PlateAlgebra1991} from attempts to address limitations of earlier distributed representations~\citep{MalsburgAssemblies1986,fodor1988connectionism}.  
HD/VSA forms a ``concept space'' by exploiting the geometry and algebra of high-dimensional spaces. 
The central idea is to represent information with randomly generated vectors, called \textit{hypervectors}.
Together with a set of operations on these hypervectors, HD/VSA can represent compositional structures, which, in turn, enables features such as reasoning by analogy~\citep{PlateAnalogical1994, KanervaLearning2000, GaylerIsomorphism2009, RachkovskijAnalogy2012, KleykoBees2015} and cognitive computing~\citep{EmruliAnalogical2013, RasmussenSpiking2014, hersche2023neuro}.   

While the foundational ideas of HD/VSA have been around for some time, it was only recently that it started gaining momentum both theoretically~\citep{FradyCapacity2018, thomas2021theoretical,clarkson2023capacity} and practically, attracting attention from the wider machine learning community~\citep{graves2014neural, RahimiBiosignal2019, NeubertRobotics2019, GanesanLearning2021, HerscheContinualLearn2022}.  
This interest is partially driven by the observation that the strengths of HD/VSA complement some of the limitations of current artificial neural networks~\citep{greff2020binding,smolensky2022neurocompositional}, leading to hybrid approaches that could be viewed as neuro-symbolic~\citep{hersche2023neuro}.

In face of the growing interest, there have been substantial efforts to consolidate and disseminate HD/VSA by its research community. 
The challenge for these initiatives has been the fact that HD/VSA involves many disciplines, including machine learning, neuroscience, electrical engineering, artificial intelligence, mathematics, and cognitive science.
Keeping up with the latest advances is not trivial as publications are spread across different venues and disciplines. 
It is, therefore, important to pursue consolidation and integration efforts, which will facilitate further research endeavors. 
Despite detailed surveys of various aspects of the area~\citep{ge2020classification, schlegel2022comparison, kleyko2021survey1, kleyko2021survey2}, much less attention, however, has been paid towards streamlining the translation of concepts, methodologies, and algorithms into a coherent open source software.

To address this issue, we developed \libraryname{}---an open source library for HD/VSA. 
Up to date with the latest advances in the area, \libraryname{} aims to lower the barrier of entry to HD/VSA for novices and provides a high-performance execution platform for experienced researchers.
To support a wide variety of HD/VSA primitives, applications, and research directions; the provided abstractions are designed to be modular, allowing users to adopt the aspects that support their workflow.
Our design philosophy focuses first on ease-of-use, aiming to be accessible without limiting expressiveness.
The ease of use allows for application-driven research to focus on the conceptual and methodological aspects of the project while, at the same time, \libraryname{} enables the study of individual elements of HD/VSA for theoretical research.
The second goal of the library is to provide high-performance execution, enabling much faster evaluation of novel algorithms without hurting the Python idiomatic development experience. 
We believe that \libraryname{} will benefit the HD/VSA and the broader machine learning communities based on its following features:
1) state-of-the-art HD/VSA methods for transforming data into hypervector;
2) support for automatic differentiation for research on hybrid neuro-symbolic models;
3) Python idiomatic interface design for a flexible, developer-first experience;
4) high-performance execution, which means that applications can run orders of magnitude faster;
5) interfaces for data sets commonly used in the literature to evaluate and benchmark HD/VSA methods (129 data sets currently).

\section{\libraryname}
\label{sec:library}

\libraryname{} builds on PyTorch~\citep{paszke2019pytorch}, a library for high-performance tensor computation. 
PyTorch and its ecosystem provide many machine learning utilities, support for various hardware accelerators, and built-in automatic differentiation (autodiff) \citep{paszke2017automatic}. 
Building on PyTorch enables the exchange of knowledge between HD/VSA and the wider machine learning community, and facilitates the cross-fertilization of ideas. 

\libraryname{} supports six common HD/VSA models, namely binary spatter codes (BSC) \citep{kanerva1997fully}, multiply-add-permute (MAP) \citep{GaylerMAP1998}, holographic reduced representations (HRR), Fourier HRR (FHRR) \citep{plate1995holographic}, sparse block codes (SBC) \citep{LaihoSparse2015,FradySDR2021}, and vector-derived transformation binding (VTB) \citep{gosmann2019vector}. 
These models differ in their specification of the high-dimensional space and their implementations of the fundamental operations: superposition and binding of hypervectors.
Support for (F)HRR enabled the implementation of vector function architectures \citep{fradyfunctions2021, frady2022computing} which extends an approach for implementing large-scale kernel machines via random Fourier features \citep{rahimi2007random}.
Our implementations are verified with extensive automatic unit testing. 
We also place emphasis on clear documentation, which is available at: 
\url{https://torchhd.readthedocs.io}

The library has six modules which provide the following functionality:
\vspace{-0.5em}%
\begin{itemize}
    \setlength\itemsep{-0.1em}
    
    \item \texttt{functional}: functions for generating hypervectors \citep{rachkovskiy2005sparse, nunes2022extension}; the operations on hypervectors \citep{schlegel2022comparison}; and a resonator network for factorizing hypervectors \citep{frady2020resonator, renner2022neuromorphic, kleyko2022integer, langenegger2023memory}.
    
    \item \texttt{embeddings}: classes for transforming scalars or feature vectors into hypervectors, ranging from simple hypervector lookups till similarity-preserving transformations of feature vectors \citep{kleyko2020density,thomas2021theoretical}, some of which approximate well-known kernels~\citep{rahimi2007random, frady2022computing}.  
    
    \item \texttt{models}: common classification models such as the centroid model with class prototypes and its various training algorithms \citep{rahimi2016hyperdimensional, imani2017voicehd, hernandez2021onlinehd, nunes2022graphhd}, learning vector quantization \citep{diao2021generalized}, and regularized least squares  \citep{kleyko2020density}.
    
    \item \texttt{memory}: methods for long-term storage and retrieval of hypervectors, such as sparse distributed memory \citep{kanerva1988sparse}, inspired by the cerebellum \citep{teeters2022separating}, the (modern) Hopfield network \citep{hopfield1982neural, krotov2016dense}, and the attention mechanism \citep{vaswani2017attention, ramsauer2020hopfield}.
    
    \item \texttt{structures}: classes for HD/VSA data structures \citep{kleyko2021vector} such as hash tables, graphs, binary trees, and finite state automata \citep{osipov2017associative, yerxa2018hyperdimensional, heddes2022hyperdimensional}. This module facilitates the development of classical algorithms using HD/VSA. 
    
    \item \texttt{datasets}: convenient access to 126 classification and 3 regression data sets commonly used in the literature. This includes a classification benchmark of UCI machine learning repository data sets \citep{Dua:2019} created by \citet{fernandez2014we}. All data sets are interoperable with the PyTorch ecosystem.
\end{itemize}

\begin{table}[tb]
    \centering
    \footnotesize
    \setlength{\tabcolsep}{5pt}
    \begin{tabular}{l|cccccccccc}
         \toprule
         Library & BSC & MAP & HRR & FHRR & \makecell{Memory\\models} & \makecell{Data\\structures} & \makecell{Resonator\\ Network} & \makecell{Auto-\\diff} & Data sets \\
         \midrule
         OpenHD & \xmark & \cmark & \xmark & \xmark & \xmark & \xmark & \xmark & \xmark & \xmark\\
         HDTorch & \cmark & \cmark & \xmark & \xmark & \xmark & \xmark & \xmark & $\bm{\pm}$ & \xmark\\
         VSA Toolbox & \cmark & \cmark & \cmark & \cmark & \xmark & \xmark & \xmark & \xmark & \xmark \\
         \libraryname{} (ours) & \cmark & \cmark & \cmark & \cmark & \cmark & \cmark & \cmark & \cmark & \cmark \\ 
         \bottomrule
    \end{tabular}
    \caption{\label{tab:lib-compare}A qualitative assessment of the functionality of HD/VSA software.}
\end{table}

We note that several other software for HD/VSA exist.
Table~\ref{tab:lib-compare} contrasts the features of \libraryname{} with those of OpenHD \citep{kang2022openhd}, HDTorch \citep{simon2022hdtorch}, and VSA Toolbox \citep{schlegel2022comparison}. 
As part of their review on HD/VSA models, \cite{schlegel2022comparison} implement a broader range of HD/VSA models, however, their VSA Toolbox misses many of the features that make \libraryname{} particularly useful for research and development of HD/VSA, see, for example, Table~\ref{tab:lib-compare}.
Importantly, \libraryname{} is the first general software library for HD/VSA since the other software were developed for more specific purposes.

To demonstrate the performance of \libraryname{}, we implemented three classification tasks and compared the execution time to the original publicly available code. 
All the experiments ran on a machine with 20 Intel Xeon Silver 4114 CPUs, 93~GB of RAM, and 4 Nvidia TITAN Xp GPUs, only a single CPU or GPU was used. 
The results in Table~\ref{tab:performance} show that \libraryname{} is faster for all examples with an average of 24$\times$ and 54$\times$ faster for CPU and GPU, respectively.
The provided abstractions ensure performant code by avoiding overly sequential execution, reinforced by our support for batch processing.

\begin{table}[h]
  \centering
  \footnotesize
  \begin{tabular}{l|ccc}
    \toprule
    Paper  & Original & CPU & GPU\\
    \midrule
    EU languages~\citep{rahimi2016robust} & 13111.73 & 542.83 (24$\times$) & \textbf{125.89} (104$\times$)\\
    EMG gestures~\citep{rahimi2016hyperdimensional} & 1152.32 & \textbf{28.44} (41$\times$) & 28.95 (40$\times$)\\
    VoiceHD~\citep{imani2017voicehd} & 277.97 & 37.74 (7$\times$) & \textbf{16.17} (17$\times$)\\
    \bottomrule
\end{tabular}
  \caption{\label{tab:performance} Execution time in seconds (and speedup)}
\end{table}
\section{Conclusions and Future Work}
\label{sec:conclusion}

\libraryname{} is a Python library for hyperdimensional computing (HD) a.k.a. vector symbolic architectures (VSA) that aims to be simple, versatile, and highly performant.
The library builds on PyTorch, thus, enabling the exchange of knowledge between the HD/VSA and the broader machine learning communities as well as making PyTorch's vast ecosystem of deep learning tools available for HD/VSA. 
\libraryname{} is designed to include any HD/VSA model and already supports binary spatter codes, multiply-add-permute, (Fourier) holographic reduced representations, sparse block codes, and vector-derived transformation binding. 
We plan to keep the library up-to-date with novel developments in the area by including more HD/VSA models and supporting a broader spectrum of learning algorithms, such as differentiable learning, regression, density estimation, and clustering. 
Our work joins the community’s vital effort to consolidate and disseminate HD/VSA research.

\newpage
\acks{%
We would like to thank Ross W. Gayler for helpful discussions on the design choices and desired functionality of the library;
Rishikanth Chandrasekaran, Dheyay Desai, Jenny Lee, and Xiaofan Yu for their contributions to the library; and finally, Anthony Thomas for his support while resolving an issue with the random projection implementation.
\newline
DK received funding from the European Union's Horizon 2020 research and innovation programme under the Marie Skłodowska-Curie grant agreement No 839179.
}

\vskip 0.2in
\bibliography{references}

\end{document}